\documentclass[10pt,twocolumn,letterpaper]{article}

\usepackage[pagenumbers]{iccv} 

\usepackage{amssymb}

\definecolor{iccvblue}{rgb}{0.21,0.49,0.74}
\usepackage[pagebackref,breaklinks,colorlinks,allcolors=iccvblue]{hyperref}

\def\paperID{9579} 
\def\confName{ICCV}
\def\confYear{2025}

\title{U-Motion: Learned Point Cloud Video Compression with U-Structured Temporal Context Generation}

\author{Tingyu Fan$^1$ \qquad Yueyu Hu$^1$ \qquad Ran Gong$^1$ \qquad Yao Wang$^1$\\
$^1$Tandon School of Engineering, New York University\\
{\tt\small \{tf2387,\quad yyhu,\quad rg4827,\quad yaowang\}@nyu.edu}
}

\begin{document}
\maketitle
\begin{abstract}
Point cloud video (PCV) is a versatile 3D representation of dynamic scenes with emerging applications. This paper introduces U-Motion, a learning-based compression scheme for both PCV geometry and attributes. We propose a U-Structured inter-frame prediction framework, U-Inter, which performs explicit motion estimation and compensation (ME/MC)  at different scales with varying levels of detail. It integrates Top-Down (Fine-to-Coarse)  Motion Propagation, Bottom-Up Motion Predictive Coding and Multi-scale Group Motion Compensation to enable accurate motion estimation and efficient motion compression at each scale. In addition, we design a multi-scale spatial-temporal predictive coding module to capture the cross-scale spatial redundancy remaining after U-Inter prediction. We conduct experiments following the MPEG Common Test Condition for dense dynamic point clouds and demonstrate that U-Motion can achieve significant gains over MPEG G-PCC-GesTM v3.0 and recently published learning-based methods for both geometry and attribute compression.
\end{abstract}    
\section{Introduction}
\label{sec:intro}







Point cloud video is a  versatile 3D representation with broad applications, including immersive volumetric video streaming, autonomous driving, and augmented reality~\cite{shan2024contrastive,hu2024low}. Due to the high data volume,  it is essential to compress point cloud videos efficiently for data storage and transmission. This demand leads to the successful standardization of MPEG PCC (including G-PCC~\cite{gpccwhitepaper} and V-PCC~\cite{graziosi2020overview}). Beyond these standards, numerous learning-based methods~\cite{wang2022sparse,sheng2021deep,zhang2023yoga} have been proposed for  compression of either static or dynamic point clouds, either for geometry (locations of points) or attributes (colors or other properties associated with points).

One essential element in point cloud video compression is inter-frame prediction, which exploits the temporal redundancy between consecutive frames. This is typically accomplished through motion estimation and compensation. However, compared to 2D videos on a regular 2D grid, the points in a PCV are irregularly spread over a small fraction of the 3D space, making  the design of motion estimation and compensation approaches for PCV significantly more challenging. 
Moreover, to ensure superior rate-distortion performance, we need to incorporate  efficient context-based coding schemes to compress the estimated motion fields as well as features for recovering the geometry or attributes.
Furthermore, it is desirable to develop a codec framework that can be used for both the geometry and attributes,  to save hardware design resources for future chip tape-outs. 
.

To address the inter-coding problem, standard approaches either project the point cloud to a 2D video and use an existing video codec for motion estimation and compensation (\textit{i.e.} V-PCC~\cite{graziosi2020overview}), or conduct cube-level motion compensation with a neighbor search strategy (\textit{i.e.} G-PCC~\cite{gpccwhitepaper}). 
Meanwhile, some learned dynamic point cloud compression methods~\cite{wang2023dynamic,wang2024versatile1,wang2024versatile2} have been proposed to leverage sparse convolution techniques~\cite{choy20194d} to exploit temporal correlation, known as {\it temporal convolution}. 
Specifically, the state-of-the-art scheme Unicorn~\cite{wang2024versatile1,wang2024versatile2} incorporates the spatial-temporal \textit{target convolution} to predict the current frame from the reference frame.  
However, these techniques limit the potential underlying motion patterns to be within the expressibility of a set of learned convolutional kernels, without the correct localization for temporal context mining offered by explicit motion estimation. This leads to suboptimal inter prediction.

To tackle the challenge of motion estimation in point cloud videos, we propose a novel unified point cloud compression scheme, named \textit{U-Motion}, which incorporates an U-Structured inter prediction module, named \textit{U-Inter}, to estimate and compensate motion at different scales. 
The U-Inter structure, akin to the popular U-Net, through the use of both top-down motion propagation, bottom-up motion predictive coding and skip connections, leverages the motion information at both upper and lower scales to inform motion estimation and coding at a current scale. U-Inter also incorporates a multi-scale group motion compensation scheme, utilizing the hierarchically reconstructed motion flow to enrich temporal context generation.

We evaluate our method following the MPEG Common Test Conditions (CTC), dense dynamic part's specification~\cite{m66563} for data splitting, in both geometry and color coding. The experimental results show that U-Motion outperforms the learning-based state-of-the-art Unicorn~\cite{wang2024versatile2}, as well as the rule-based MPEG standard G-PCC-GesTM-v3.0~\cite{w23041} by a large margin in terms of rate-distortion (R-D) trade-off. We conduct ablation studies to validate the effectiveness of each component in our proposed U-Motion codec. Our contributions can be summarized as follows:
\begin{itemize}
    \item We propose a novel unified PCV compression scheme, named U-Motion, which incorporates a hierarchical {\bf U-Inter} Prediction module using a U-Net like structure. The proposed U-Inter incorporates {\bf Top-Down Motion Propagation}, {\bf Bottom-Up Motion Predictive Coding} and {\bf Multi-scale Group Motion Compensation} for accurate motion generation and efficient motion coding under rate constraint. The framework can be used for both geometry and attribute coding.
    \item We evaluate the  U-Motion codec following the MPEG CTC's data splitting, for both geometry coding and color coding, and demonstrate  U-Motion's significant performance gain against both the  state-of-the-art learning-based method Unicorn \cite{wang2024versatile2} and MPEG standard G-PCC-GesTM v3.0  \cite{w23041}, in terms of rate-distortion trade-off.
\end{itemize}
\section{Related Work}
\subsection{Traditional Point Cloud Video Compression} 
There are two primary technical routes for traditional point cloud video compression: 3D-structure-based and video-based. 
3D-structure-based methods~\cite{thanou2015graph, de2017motion,gpccwhitepaper} process original point clouds directly and leverage various block-matching algorithms to identify temporal dependencies. 
A notable example is the MPEG standard G-PCC-GesTM~\cite{w23041}, which has demonstrated state-of-the-art performance among all 3D-structure-based methods. 
In contrast, 2D-video-based methods~\cite{zhu2020view,he2017best,graziosi2020overview} reuse the existing 2D video codecs by projecting the 3D point cloud video into 2D space.
Among all 2D-video-based methods, the MPEG standard V-PCC achieves state-of-the-art performance.

\begin{figure}[t]
  \centering
   \includegraphics[width=0.99\linewidth]{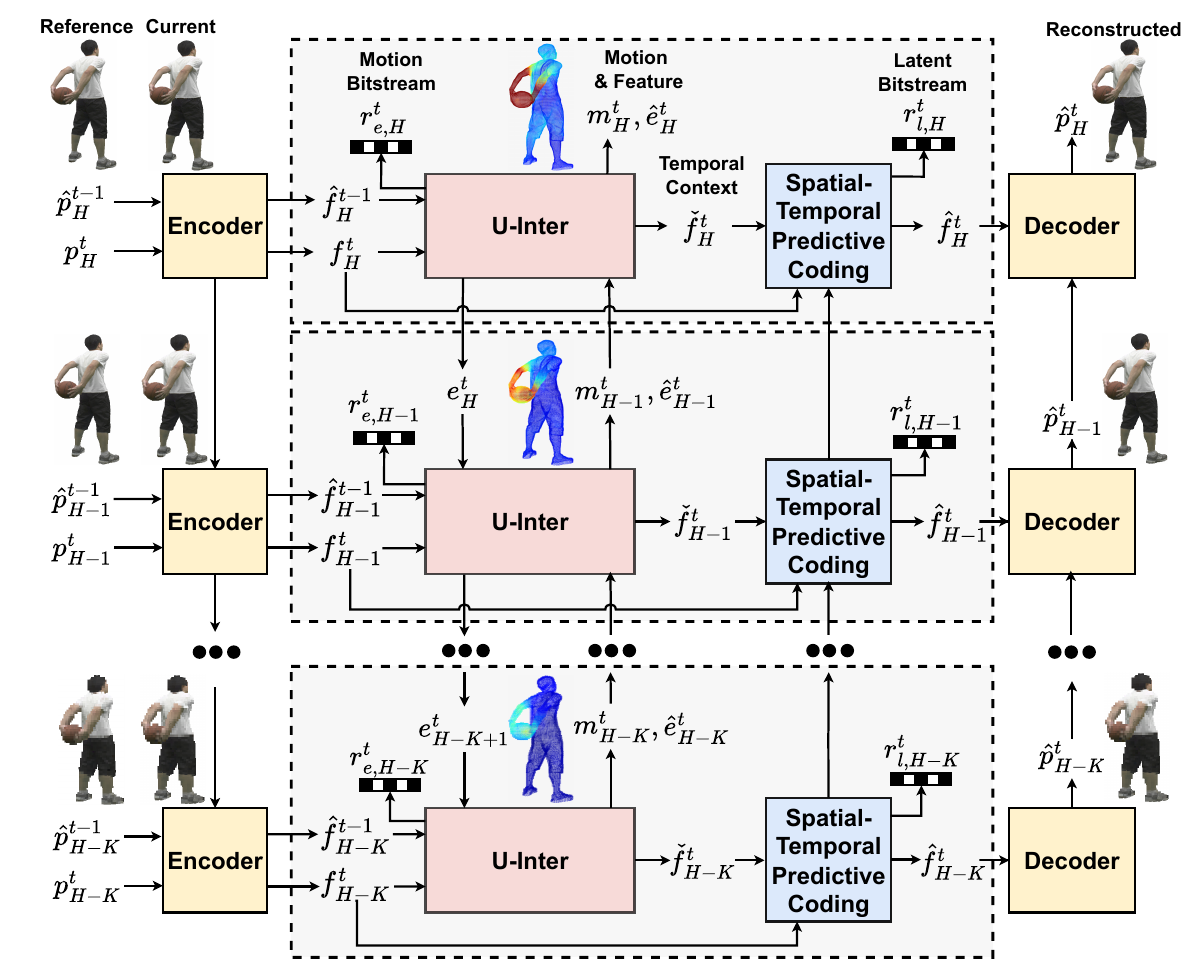}

   \caption{The overall architecture of U-Motion for attribute. }
   \label{fig:overall}
\end{figure}

\subsection{Learning-Based Point Cloud Video Compression}
Early research on learning-based point cloud compression focuses on static point clouds~\cite{wang2021multiscale,wang2022sparse,fang20223dac,zhang2023yoga,fu2022octattention}. Wang \textit{et al.}~\cite{wang2021multiscale} first propose to use sparse convolution~\cite{choy20194d} to compress voxelized point clouds, and propose a multiscale architecture, as well as a local auto-regressive architecture SOPA~\cite{wang2022sparse} for lossless geometry compression.  To further exploit the temporal redundancy in point cloud video, some methods adopt temporal convolution for inter prediction. Anique \textit{et al.}~\cite{akhtar2024inter} design a predictor for geometry compression, which compensates and compresses the current frame's latent representation based on its reference neighbors. Wang \textit{et al.} introduce Unicorn~\cite{wang2024versatile1,wang2024versatile2}, a unified architecture that integrates geometry \& attributes, static \& dynamic compression, using context compensation and multiscale motion compensation. The motion compensation is obtained through sparse convolution on the nearby voxels in the previous frame, known as target convolution. However, both Unicorn and Anique \textit{et al.} lack explicit motion estimation, leading to sub-optimal temporal prediction. On the contrary, Fan \textit{et al.} propose D-DPCC~\cite{ijcai2022p126} with an end-to-end learning-based motion estimation and motion compensation (ME/MC) module. Xia \textit{et al.}~\cite{xia2023learning} further propose a two-layer ME/MC architecture for coarse-to-fine motion compensation. Jiang \textit{et al.}~\cite{jiang2023end} introduces multiscale motion estimation, where finer-scale's motion estimation uses coarser scale's estimated motion as base motion.  However, the above ME/MC-based methods are developed only for point cloud video geometry. In addition, their lack of a hierarchical inter- and intra-prediction structure leads to inferior coding efficiency compared to Unicorn-Geometry.

\subsection{Learning-based Video Compression}
Prior to learning-based point cloud compression, there have been significant advances in learned video compression~\cite{lu2019dvc,hu2021fvc,hu2022coarse,li2021deep,sheng2022temporal,li2022hybrid,li2023neural,li2024neural}. DVC~\cite{lu2019dvc} and FVC~\cite{hu2021fvc} are the pioneering AI video codecs that perform ME/MC on pixel domain and feature domain, respectively. These works follow a residual-coding framework which encodes the difference between the warped decoded reference frame and the current frame.
Li \textit{et al.} propose DCVC~\cite{li2021deep}, which extends the residual-coding framework to context-coding,  using motion-compensated reference frame features as temporal context for coding the current frame features.
DCVC-HEM~\cite{li2022hybrid} further proposes a parallel checkerboard entropy model supporting variable bit-rate. DCVC-DC~\cite{li2023neural} and DCVC-FM~\cite{li2024neural} further incorporate hierarchical quality structure and inter feature modulation to alleviate the error accumulation  in the temporal context. However, due to the irregular point geometry  and data diversity of point cloud sequences, learning-based video compression methods cannot be directly integrated to point cloud video compression.
\section{Methodology}\label{method}

\subsection{Overview of Attribute Compression}

As shown in Figure \ref{fig:overall}, U-Motion is a hierarchical architecture consisting of $K+1$ ($K=4$ during our experiments) U-Inter and Spatial-Temporal Predictive Coding layers, where the $l$-th layer corresponds to the $l$-th level of the point cloud when represented by an octree, with the highest layer $H$ corresponding to the original voxelized point cloud. 
We denote the point cloud frame (including both geometry coordinate and color attribute for each point at each octree level) at time $t$ and layer $l$ as $p_l^{t}$, and its reference ({\it i.e.} the decoded previous point cloud frame) as $\hat{p}_l^{t-1}$. $p_l^t$ is expressed as a sparse tensor $\left[G(p_l^t),A(p_l^t)\right]$, where $G(p_l^t)$ denotes an $N\times3$ geometry coordinate matrix and $A(p_l^t)$ denotes the corresponding $N_l\times3$ attribute matrix, with $N_l$ denoting the total number of points at layer $l$. The encoder  fuses $p_l^{t}$ and the upper layer's latent  $f_{l+1}^{t}$ (after down-sample) to  generate the latent representation $f_l^{t}$. The same encoder is also used to generate the reference frame's latent $\hat{f}_l^{t-1}$.

The  U-Inter module takes the latents $f_l^t$ and $\hat{f}_l^{t-1}$, as well as the upper-layer  motion feature $e_{l+1}^t$ (through a top-down connection) and lower-layer decoded motion features $\hat{e}_{l-1}^t$ and decoded motion $m_{l-1}^t$ (through a bottom-up connection),  and outputs the motion feature ${e}_l^t$ at layer $l$. Furthermore, it compresses ${e}_l^t$ to generate the motion bit stream $r_{e,l}^t$ as well as decoded motion feature $\hat{e}_l^t$.  Thereafter, it decodes the motion field $m_l^t$ from $\hat{e}_l^t$ and generates the temporal context $\check{f}_l^t$, by warping $\hat{f}_l^{t-1}$ using  ${m}_l^t$.

The Spatial-Temporal Predictive Coding module encodes and decodes current frame's latent $f_l^t$ into $\hat{f}_l^t$ based on the temporal context $\check{f}_l^t$ and spatial context  $\hat{f}_{l-1}^t$. Finally, the decoder module takes $\hat{f}_{l}^{t}$ and the lower layer's decoded points $\hat{p}_{l-1}^{t}$ to recover the color attributes in $\hat{p}_l^{t}$. 

\subsection{Hierarchical Attribute Encoding and Decoding}
\paragraph{Latent Feature Encoder.} The encoder in Fig. \ref{fig:overall} maps the current layer $p_l^{t}$ and the upper-layer latent $f_{l+1}^{t}$ into the latent representation for the current layer $f_l^{t}$. The same encoder is used to map   $\hat{p}_l^{t-1}$ and $\hat{f}_{l+1}^{t-1}$ into $\hat{f}_l^{t-1}$. In contrast to Unicorn, which blocks the latents from flowing between layers, our encoder uses both $p_l^{t}$ and $f_{l+1}^{t}$ to generate $f_l^{t}$, so that the coarser layer's latent encapsulates important information of the finer layers.
We use sparse convolution for network construction, as with all other modules of the entire network (See supplementary).

\begin{figure}[t]
  \centering
   \includegraphics[width=1.0\linewidth]{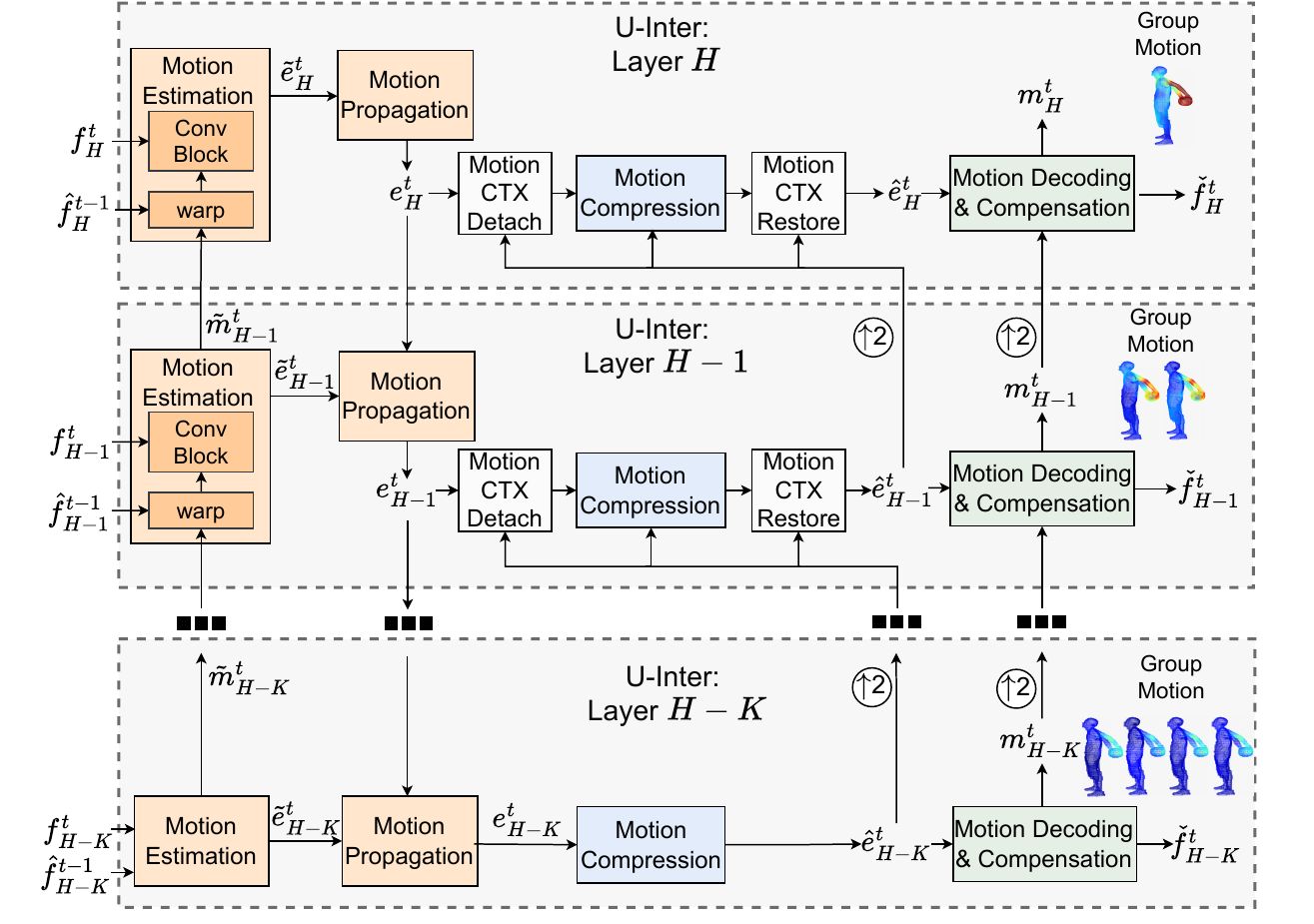}

   \caption{The network architecture for U-Inter module.}
   \label{fig:uinter}
\end{figure}

\paragraph{Attribute Decoder.} We assume the geometry has been decoded losslessly and is known. The decoder module recovers the colors in the $l$-th scale $\hat{p}_l^t$ based on the reconstructed latent $\hat{f}_l^t$ and coarser scale's reconstruction $\hat{p}_{l-1}^t$. Specifically, the network estimates a residual attribute $\Delta p_l^t$ at every layer based on the concatenation of upsampled $\hat{p}_{l-1}^t$ and $\hat{f}_l^t$. The reconstruction is the sum of $\Delta p_l^t$ and the upsampled $\hat{p}_{l-1}^t$. The upsample is realized by trans-pooling.

\subsection{U-Inter}



\paragraph{U-Structured Inter Prediction.} As shown in Figure \ref{fig:uinter}, our U-Inter Prediction follows a U-Net structure, consisting of bottom-up (coarse-to-fine) motion estimation, top-down motion feature propagation,  bottom-up predictive coding of motion, and finally multi-scale motion compensation (warping based on the decoded motions at different scales).

\begin{figure}[t]
  \centering
   \includegraphics[width=0.99\linewidth]{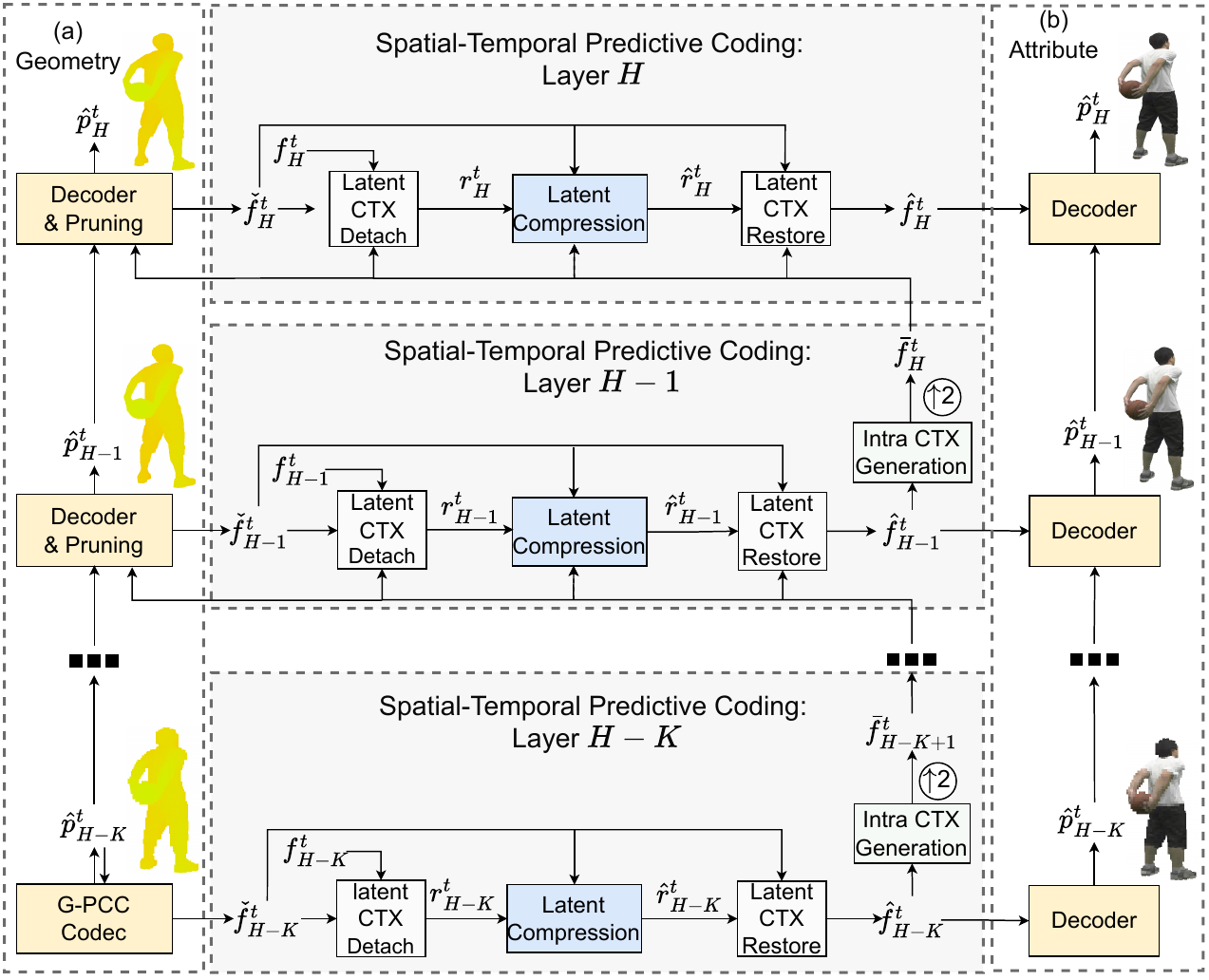}

   \caption{The network architecture for Spatial-Temporal Predictive Coding. For attribute compression, (b) is used for reconstruction; for geometry compression, (a) is used for reconstruction.}
   \label{fig:spc}
   \vspace{-2mm}
\end{figure}

Specifically, we adopt the coarse-to-fine motion estimation widely used in video/point cloud compression \cite{xia2023learning,jiang2023end}. For layer $l$, the network first warps reference $\hat{f}_l^{t-1}$ based on the coarse encoder motion $\tilde{m}_{l-1}^t$. 
The convolution block performs spatial convolution on the concatenated feature channels from both warped $\hat{f}_l^{t-1}$ and ${f}_l^{t}$  (using sparse convolution),
and outputs encoder motion feature $\tilde{e}_{l}^t$, which encodes the residual motion at layer $l$.  This block also decodes $\tilde{e}_{l}^t$ into $\tilde{m}_l^t$ for the motion estimation at the upper scale.


Instead of compressing  $\tilde{e}_l^t$ directly, we propose a  motion propagation block that fuses $\tilde{e}_l^t$ with $e_{l+1}^t$ (through a top-down connection), and generates fused motion feature $e_l^t$, which contains motion information of layers $l:H$. This block enables  1) propagation of  high-frequency details to the coarser scale, important for correct motion estimation at each scale; 2) distribution of finer-scale motion across multiple scales, to facilitate more efficient and flexible compression of motion information. In later ablation studies, we will show that when this propagation block is  removed, the network often yields degraded motions in part due to the rate constraint. Note that this top-down motion propagation introduces redundancies among $e_l^t$ across layers. Therefore, the bottom-up motion prediction module  \textit{contextually detaches} the redundancy in $e_{l}^{t}$ that is predictable from the decoded coarser-scale $\hat{e}_{l-1}^{t}$, when compressing $e_{l}^{t}$, and \textit{contextually restores} $\hat{e}_{l-1}^{t}$ into $\hat{e}_l^t$ on the decoder side.

\paragraph{Multiscale Motion Decoding and Compensation.} Although \cite{jiang2023end} introduces coarse-to-fine motion estimation, it only applies motion compensation at the finest scale to code the latent features using temporal prediction at the finest scale.
We propose  progressive coarse-to-fine motion compensation with multiples scales of compensation and predictive coding. 
The motion decoding module decodes a residual motion $\Delta m_{l}^{t}$ from motion feature $\hat{e}_l^t$, which is then added to the upsampled $m_{l-1}^{t}$ to generate the motion  $m_{l}^{t}$.

\paragraph{Group Motion}
As shown in \cite{li2023neural} for motion estimation in 2D video, different feature channels may experience different motions. Borrowing this idea, we  estimate multiple motion groups at each scale.
The motion decoding module  outputs a motion $m_l^t\in \mathbb{R}^{N_l^t\times G_l\times 3}$, where $N_l^t$ is the number of points in frame $t$ layer $l$, $G_l$ is the number of motion groups of layer $l$. Assuming the latent dimension is $C_l$, we divide the  $C_l$ feature channels evenly into $G_l$ groups, and decode from the motion features $\hat{e}_l^t$ a residual motion flow for each group, $\Delta m_{l,g}^t$.  
In general, we find that increasing the number of groups brings gain, despite the potential motion bit rate increase brought by coding more motion groups. However, increasing the group size results in a quadratic increase in complexity for motion compensation. Consequently, using large motion groups at fine scales leads to a sharp rise in computational complexity. Therefore, we employ motion group merging, utilizing large group sizes at coarse scales and progressively merging and reducing the number of motion groups at finer scales.

To derive the decoded motion at scale $l$ for group $g$, we need to generate the upsampled motion from scale $l-1$, which generally has more groups. We average the upsampled motions for the corresponding groups at scale $l-1$, and add to the the decoded residual motion. That is: 

\begin{equation}
    m^t_{l,g}=\frac{1}{G'}\sum_{i=G'g}^{G' (g+1)-1}\uparrow m_{l-1,i}^{t}+\Delta m_{l,g}^t,
\end{equation}
where $G'=\lceil \frac{G_{l-1}}{G_l}\rceil$, $\uparrow$ denotes the upsample operator

For each point with coordinate $s$ in $f_l^t$, its warped coordinate for channel $c$, $\check{s}_c=s+m_{l,s,g}^t$,
may not have a correspondence  in $\hat{f}_{l}^{t-1}$, where $g$ denotes the group number corresponding to channel $c$, $m_{l,s,g}^t\in\mathbb{R}^{3}$ denotes the motion at time $t$, layer $l$, coordinate $s$ and group $g$. Therefore, we adopt the 3D Adaptive Weighted Interpolation (3DAWI) algorithm \cite{ijcai2022p126} to generate the inter context $\check{f}_{l}^{t-1}$,
\begin{equation}
    \check{f}_{l,s_c}^{t}=\frac{\sum_{v\in \vartheta(\check{s}_c)}d_{v,\check{s}_c}^{-1}\cdot \hat{f}_{l,v}^{t-1}}{\max\left\{\sum_{v\in \vartheta(\check{s}_c)}d_{v,\check{s}_c}^{-1},\alpha\right\}},
\end{equation}
where $\vartheta(\check{s}_c)$ denotes the K-nearest neighbors of $\check{s}_c$ in $\hat{f}_l^{t-1}$, $d_{v,\check{s}_c}$ is the distance between $\check{s}_c$ and $v$, $\alpha$ is a penalty parameter for points too far from $\check{s}_c$.

\subsection{Spatial-Temporal Predictive Coding}\label{chapter:msintra}

\paragraph{Layer-wise redundancy.} Besides inter-frame temporal redundancy, $f_l^t$ exhibits inter-layer spatial redundancy. 
Therefore, as shown in Figure \ref{fig:spc}, Spatial-Temporal Predictive Coding module further removes the spatial redundancy by estimating an intra context $\bar{f}_l^t$, using the already decoded lower-scale latent  $\hat{f}_{l-1}^t$.

\begin{figure*}[htbp]
  \centering
   \includegraphics[width=0.95\linewidth]{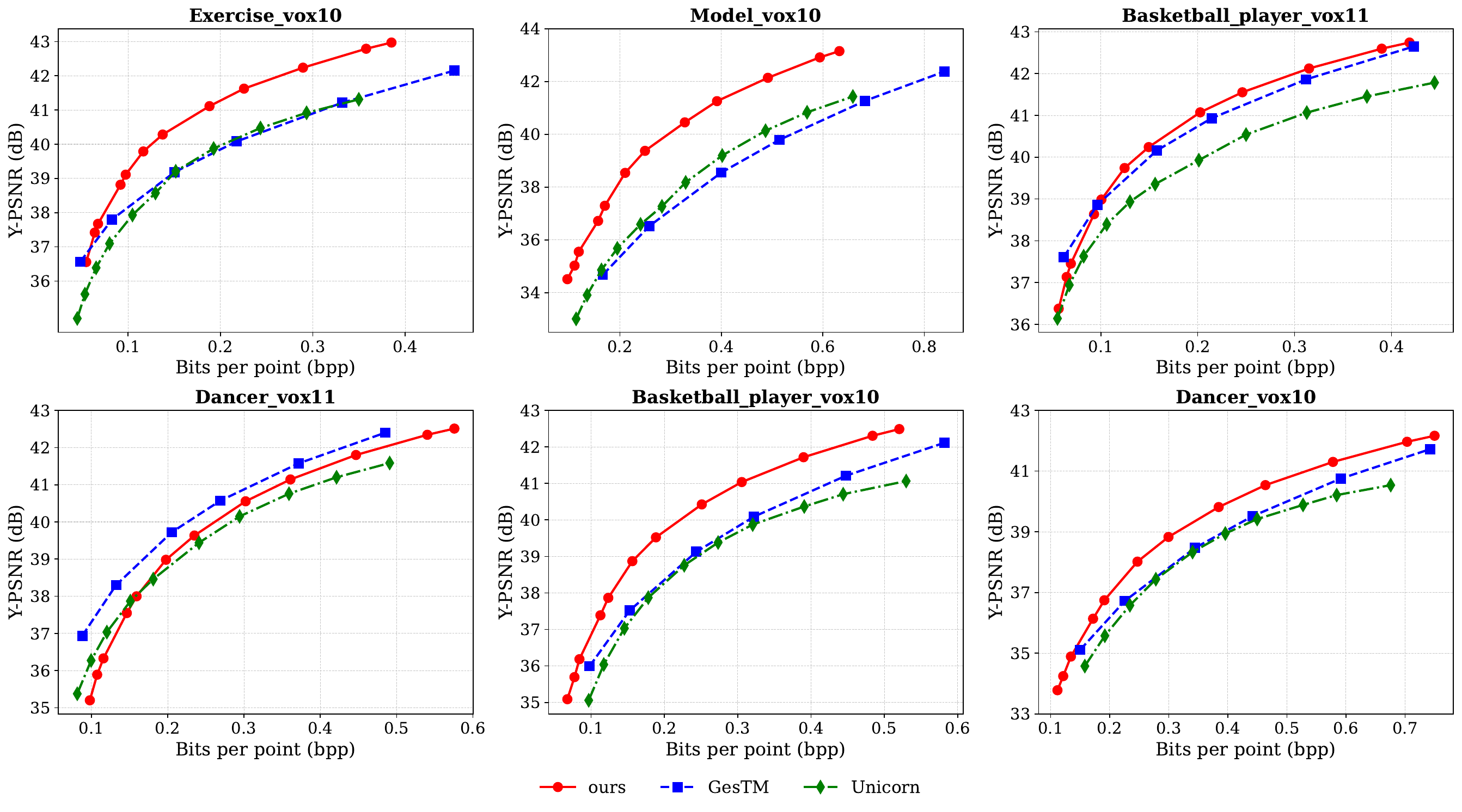}

   \caption{Y-PSNR Performance comparison on attribute (color) compression among our method, Unicorn and G-PCC-GesTM.}
   \vspace{-2mm}
   \label{fig:res_attr_y}
\end{figure*}

\begin{table}[t]
    \centering
    \small
     \caption{Attribute compression BD-Rate(\%) gains against G-PCC-GesTM and Unicorn. Negative values indicate bit rate reduction under same quality. \textbf{CTC-Overall} denotes the average test result on MPEG CTC sequences (The first four sequences). \textbf{Overall} denotes the average test result on all six sequences.}
    \begin{tabular}{cc cc cc}
      \toprule
      \multicolumn{2}{c}{Sequence} & \multicolumn{2}{c}{G-PCC-GesTM} & \multicolumn{2}{c}{Unicorn}  \\ 
      \multicolumn{2}{c}{ } & Y & YUV & Y & YUV\\ 
      \midrule
      \multicolumn{2}{c}{exercise\_vox10} & -29.69  & -25.02  & -33.75  & -37.98\\
      \multicolumn{2}{c}{model\_vox10} & -43.62  & -41.87  & -38.32  & -43.63\\
      \multicolumn{2}{c}{basketball\_vox11} & -4.05  & 1.89  & -24.38  & -32.50\\
      \multicolumn{2}{c}{dancer\_vox11} & 20.92  & 28.94  & 0.04  & -10.14\\
      \multicolumn{2}{c}{basketball\_vox10} & -26.72  & -22.07  & -32.40  & -37.72\\
      \multicolumn{2}{c}{dancer\_vox10} & -16.43  & -11.22  & -22.05  & -31.77\\
      \midrule
      \multicolumn{2}{c}{\textbf{CTC-Overall}} & \textbf{-14.11}  & \textbf{-9.84}  & \textbf{-24.10}  & \textbf{-31.06}\\
      \multicolumn{2}{c}{\textbf{Overall}} & \textbf{-16.58}  & \textbf{-12.10}  & \textbf{-25.14}  & \textbf{-32.29}\\
      \bottomrule
     \end{tabular}
      \label{tab:bd_attr}
\end{table}

\paragraph{Context Detach and Restore.}
Given latent representation $f$ to be coded and the context $f_c$, the network generates residual $r$ from the unpredictable part of $f$ given $f_c$. Inspired by the conditional coding in \cite{li2021deep}, we propose a context detach module with residual connection:
\begin{equation}
    r=f-\phi_{enc}(f|f_c;\theta_{enc}),
\end{equation}
the corresponding context restore operation, which recovers $\hat{f}$ from the reconstructed residual $\hat{r}$ and context $f_c$, is expressed as:
\begin{equation}
    \hat{f}=\hat{r}+\phi_{dec}(\hat{r}|f_c;\theta_{dec}),
\end{equation}
where $\phi_{enc}$ and $\phi_{dec}$ are two neural networks.
The proposed context detach and restore operation grants the network with equivalent flexibility as conditional coding \cite{li2021deep}, but also provides the network with a ``shortcut" to narrow $r$'s distribution.  For motion $e_l^t$, context $f_c$ is $e_{u,l-1}^t$. For latent $f_l^t$, context $f_c$ is inter and intra context $\check{f}_l^t$ and  $\bar{f}_l^t$.

\subsection{Entropy Coding and Variable Rate Coding}
The entropy coding module encodes the input $x$ ($x$ is the feature $r$ after the context detach module, $r=e_l^t$ for motion compression and $r=f_l^t$ for latent compression) into a bitstream. We adopt the popular feature compression approach, which uses both hyperprior and  context features to estimate the probability distribution for quantized $x$ \cite{balle2017end,balle2018variational}.  See more detail in  the supplement. 
For motion compression, the context $x_c=\left[e_{u,l-1}^t, m_{l-1}^t\right]$. For latent compression, $x_c=\left[\check{f}_l^t,\bar{f}_l^t\right]$. We adopt the global-and-local quantization scheme proposed in DCVC-FM \cite{li2024neural}, which generates the quantization stepsizes from a given rate control parameter $\lambda$. This enables a variety of bit rates with only one compression model.






\begin{figure*}[t]
  \centering
   \includegraphics[width=0.98\linewidth]{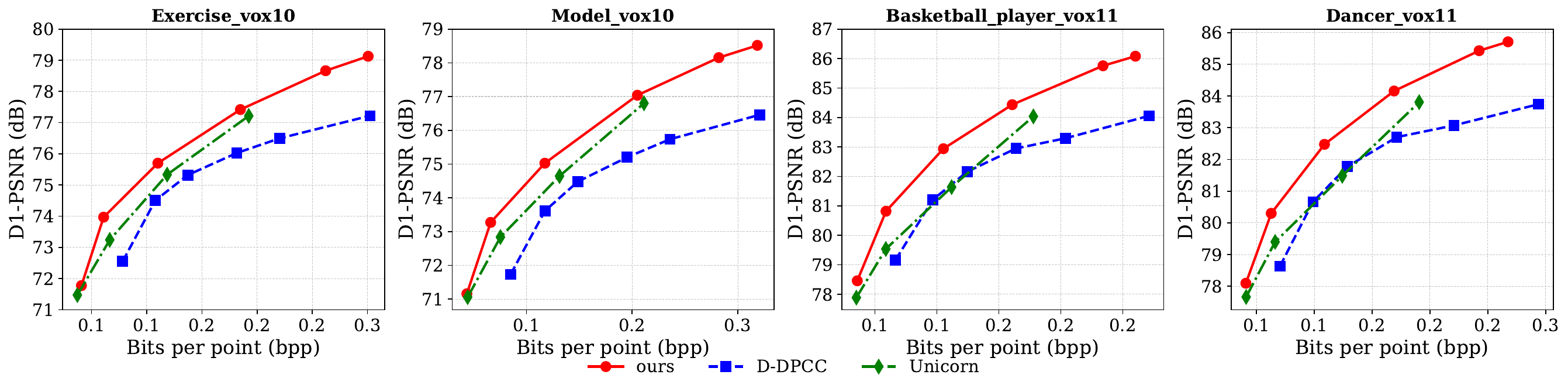}
    \vspace{-2mm}
   \caption{D1-PSNR performance comparison on lossy geometry compression among our method, Unicorn and D-DPCC. The inconsistency of Unicorn and D-DPCC's RD-curve compared with that in their original paper is due to different quantization methods used when downsampling 11-bit point clouds into 10-bit. Unicorn and D-DPCC~\cite{wang2024versatile1,wang2024versatile2,ijcai2022p126} used $floor(\cdot)$ for quantization. Instead, We follow MPEG's standard that uses $round(\cdot)$ for quantization. We have confirmed this with the authors of Unicorn and D-DPCC.}
   \label{fig:res_geo_d1}
   \vspace{-2mm}
\end{figure*}

\subsection{Geometry Compression}\label{geom}
Our network can also be used for point cloud video geometry compression with slight modification. For geometry, the $l$-th layer point cloud $p_l^t=\left[G(p_l^t),O(p_l^t)\right]$, where $O(p_l^t)$ is an all-one vector indicating the occupancy. The same encoder structure is used to generate the latent $f_l^t$ and $\hat f_{l}^{t-1}$. To code $f_l^t$, because the actual geometry is not known yet at the decoder, we cannot generate the inter context $\check{f}_l^t$.  Rather we first reconstruct $\hat{p}_l^t$ from $\bar{f}_l^t$, which is generated from $\hat{f}_{l-1}^t$ by sparse generative deconvolution with no target coordinates specified. The reconstruction is done by the APU module in Unicorn~\cite{wang2024versatile1}, which will estimate the probability that each voxel in $\bar{f}_l^t$ is occupied. For lossy compression, the top-N voxels in $\bar{f}_l^t$ with the highest occupancy probability will be kept to reconstruct $\hat{p}_l^t$; for lossless compression, arithmetic coder will be used to code $p_l^t$ losslessly using the estimated occupancy probability (thus $\hat{p}_l^t=p_l^t$). Based on the decoded  $\hat{p}_l^t$, we use the same U-Inter module shown in Fig.~\ref{fig:overall} to generate  the temporal context $\check{f}_l^t$. Then using  $\check{f}_l^t$ and spatial context $\bar{f}_l^t$, we code the current frame latent $f_l^t$ using the network architecture shown in Fig.~\ref{fig:spc} (a). In other word, for geometry, ${f}_l^t$ is coded to help the reconstruction of $\hat{p}_{l+1}^t$ instead of $\hat{p}_{l}^t$. 

\begin{table}[t]
    \centering
    \small
     \caption{Geometry compression BD-Rate(\%) gains against Unicorn and D-DPCC.  \textbf{CTC-Overall} denotes the average test result on all the four sequences from MPEG CTC.}
    \begin{tabular}{cc cc cc}
      \toprule
      \multicolumn{2}{c}{Sequence} & \multicolumn{2}{c}{Unicorn} & \multicolumn{2}{c}{D-DPCC}  \\ 
      \multicolumn{2}{c}{ } & D1 & D2 & D1 & D2\\ 
      \midrule
      \multicolumn{2}{c}{exercise\_vox10} & -17.95  & -23.31  & -35.45  & -35.55\\
      \multicolumn{2}{c}{model\_vox10} & -18.05  & -24.50  & -38.95  & -36.29\\
      \multicolumn{2}{c}{basketball\_vox11} & -28.86  & -30.78  & -38.42  & -37.55\\
      \multicolumn{2}{c}{dancer\_vox11} & -26.15  & -24.38  & -36.69  & -34.35\\
      \midrule
      \multicolumn{2}{c}{\textbf{CTC-Overall}} & \textbf{-22.75}  & \textbf{-25.74}  & \textbf{-37.37}  & \textbf{-35.93}\\
      \bottomrule
     \end{tabular}
     \vspace{-2mm}
      \label{tab:bd_geo}
\end{table}  
\section{Experiments}
\subsection{Training and Testing Datasets}
We follow the MPEG AI-3DG Common Test Condition (CTC) for dense dynamic point clouds, which specifies using the 8i Voxelized Full Bodies (8iVFB)~\cite{d20178i} dataset for training, and the Owlii~\cite{keming2018owlii} dataset for testing.  8iVFB has a data precision of 10 bits and contains four sequences: {\it soldier} , {\it longdress} , {\it redandblack}  and {\it loot}. Owlii contains four sequences in 11 bits precision: {\it basketball\_player}, {\it dancer}, {\it exercise}, {\it model}. Following MPEG CTC's requirement, we use 10-bit versions of {\it model} and {\it exercise} for testing provided by MPEG.

\subsection{Experimental Settings}
\paragraph{Network \& Training Details.} For attribute compression, we use the layer-wise joint rate-distortion loss function:

\begin{equation}
L=\sum_{l=L}^H R_{r,l}+R_{m,l}+\lambda ||p_l^t-\hat{p}_l^t||_2^2,
\end{equation}
where $R_{r,l}$ is the bit-rate of latent at layer $l$ and $R_{m,l}$  the bit-rate of motion. Note that $p_l^t$ is in YUV-space. We train only one model with $\lambda \in \left[256,18000\right]$. $\lambda$ is randomly picked from $\left[256,18000\right]$ for each iteration. We use $5$ layers to construct our network, {\it i.e.} $K=4$. The highest scale $H$ equals to the precision of data (11/10 for 11/10-bit point clouds). We choose not to have a U-Inter module in the highest scale $H$ due to complexity issue, as the K-NN search in scale $H$'s motion compensation brings excessive computational cost.
We use only the spatial context $\bar{f}_H^t$ updated from $H-1$ 
 scale's decoded feature $\hat{f}_{H-1}^t$ when coding scale $H$.
We train our model with one NVIDIA A100-SXM4-80GB. The overall training time is 4 days. During training, we use the lossless previous frame, {\it i.e.} $p^t$ as reference and only train one P-frame for each iteration. For geometry compression, the training details are in supplementary material. 

\paragraph{Baselines.} For learning-based baseline, we compare with learning-based state-of-the-art Unicorn \cite{wang2024versatile1,wang2024versatile2}. Because the code for Unicorn is not publicly available, we implemented their methods for both geometry and attribute compression. We strictly follow the setting reported in the paper and train their model also for 4 days for fairness. For attribute, the results reported in \cite{wang2024versatile2} were obtained by mixing the 6 sequences from 8iVFB and Owlii for training and testing on {\it soldier} from 8iVFB and {\it basketball\_player} from Owlii. We train their model using 8iVFB dataset only to meet the MPEG CTC, which yields lower performance for {\it basketball\_player} than that reported in \cite{wang2024versatile2}  due to the data distribution difference. 
%
For geometry compression, we also compare with D-DPCC \cite{ijcai2022p126}, and the test results were provided by the authors.

For rule-based attribute compression standards, we compare with  MPEG G-PCC-GesTM-v3.0 following the settings in \cite{wang2024versatile2}. For geometry compression, as results shown in D-DPCC and Unicorn have demonstrated significant gain against G-PCC-GesTM and V-PCC, we only compare with learning-based methods D-DPCC and Unicorn. 

\vspace{-2mm}\paragraph{Metrics.} We use MPEG reference software {\it pc\_error} to calculate distortion: D1-PSNR and D2-PSNR for geometry; Y-PSNR and YUV-PSNR (See supplementary material) for attribute. We also compare the BD-rate of each R-D curve.

 \begin{figure}
    \centering
    \scriptsize
    \begin{subfigure}{0.24\linewidth}
        \includegraphics[width=\linewidth]{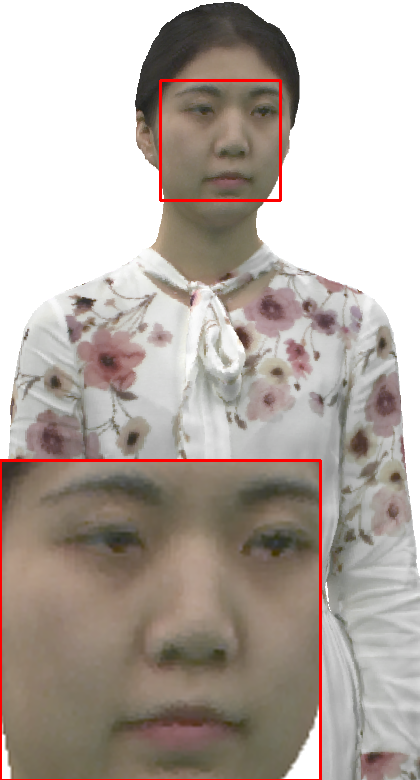}
    \end{subfigure}
    \begin{subfigure}{0.24\linewidth}
        \includegraphics[width=\linewidth]{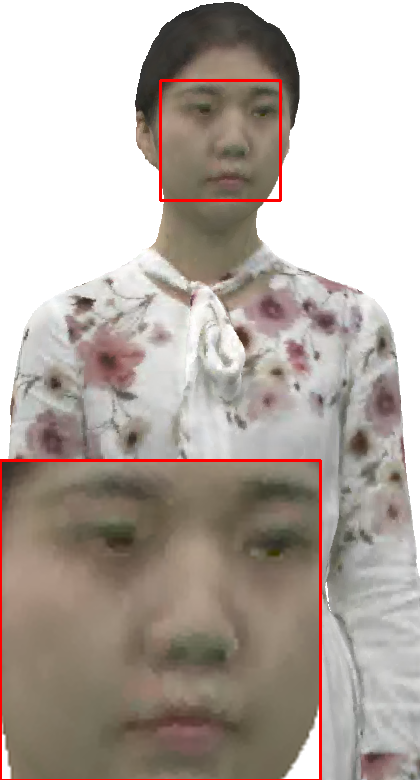}
    \end{subfigure}
    \begin{subfigure}{0.24\linewidth}
        \includegraphics[width=\linewidth]{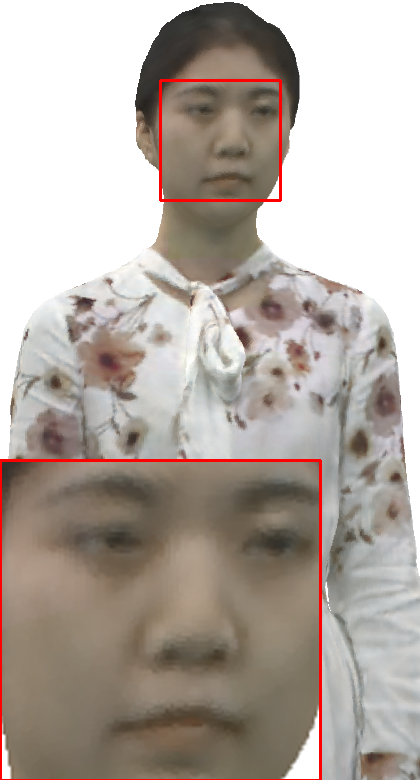}
    \end{subfigure}
    \begin{subfigure}{0.24\linewidth}
        \includegraphics[width=\linewidth]{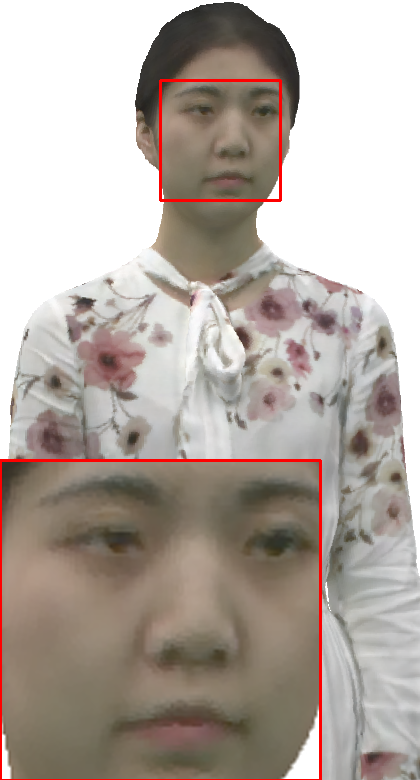}
    \end{subfigure}

    \begin{subfigure}{0.24\linewidth}
    \centering
    Original
    \end{subfigure}
    \begin{subfigure}{0.24\linewidth}
    \centering
    Ges-TM \\ (0.170 bpp)
    \end{subfigure}
    \begin{subfigure}{0.24\linewidth}
    \centering
    Unicorn \\ (0.177 bpp)
    \end{subfigure}
    \begin{subfigure}{0.24\linewidth}
    \centering
    Ours \\ (0.164 bpp)
    \end{subfigure}
    
    \caption{Rendering results from reconstructed point clouds.}
   \vspace{-2mm}
    \label{fig:recon_render}
\end{figure}

\subsection{Results}
\paragraph{Attribute Compression.} The R-D curves of attribute compression performance comparison is shown in Figure \ref{fig:res_attr_y}, with the corresponding BD-Rate detailed in Table \ref{tab:bd_attr}. Please refer to the supplementary material for R-D curves with the YUV-PSNR. We assume the point cloud geometry is losslessly coded and not factored into the bit rate consumption. In addition, to remove the influence of I-frame coding, we use the original point cloud as I-frame, and do not consider the bits of I-frame when determining the bit rate (for U-Motion and baselines).   We test the first 32 frames in each sequence.
The original first frame is used as the decoded I-frame, and each subsequent frame is coded as a P-frame using U-Motion. 
Our method achieves significant gain against G-PCC-GesTM v3.0 and Unicorn on two 10-bit MPEG sequences {\it exercise} and {\it model}. Our method does not perform as well on 11-bit sequences {\it basketball\_player} and {\it dancer}  as on 10-bit sequences. We attribute this to the fact that the MPEG-specified training data are all 10-bit and have relatively smaller motion than these two 11 bits sequences, leading to a significant difference between training and testing data. 
When compared to the 10 bit versions of these two sequences,  {\it basketball\_player\_vox10} and {\it dancer\_vox10}, our method achieve more significant gains.  Figure~\ref{fig:recon_render} shows that U-Motion achieves better visual quality at a lower bit-rate compared to other methods (sequence {\it model\_vox10}). 

\paragraph{Geometry Compression.} Figure \ref{fig:res_geo_d1} shows the lossy  geometry compression performance comparison in terms of  D1-PSNR. D2-PSNR R-D curve is in supplementary material.  Table \ref{tab:bd_geo} reports both the D1-PSNR and D2-PSNR BD-rate. It's shown that our method outperforms the previous state-of-the-art method Unicorn and the first learning-based point cloud video compression method D-DPCC.

 \begin{figure}[t]
  \centering
   \includegraphics[width=0.97\linewidth]{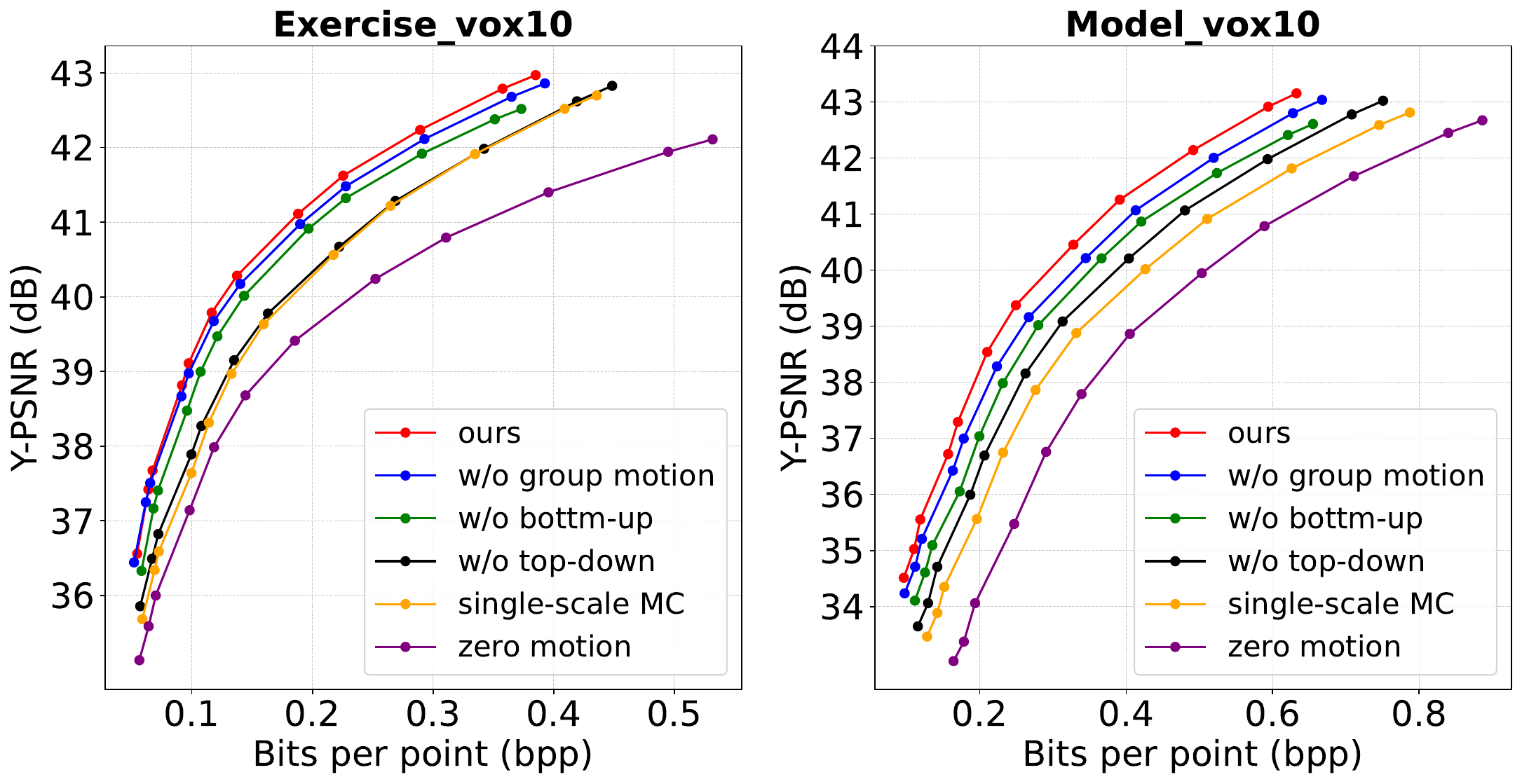}

   \caption{Ablation study on U-Inter's different modules.}
    \vspace{-2mm}
   \label{fig:abl_motion_contribution}
\end{figure}

\section{Ablation Study}
\subsection{Impact of U-Inter}  \label{chapter:abl_motion}
Figure \ref{fig:abl_motion_contribution} compares our method U-Motion with four variations, differing in motion estimation and coding. 
\vspace{-2mm} \paragraph{Gain Brought by Motion.} As the baseline codec Unicorn uses no motion information, we train  a variation of the U-motion model with ``zero motion''. The temporal context at each point is the interpolated feature of its $K$ nearest neighbors in the reference frame.
 Despite zero motion bits used, the temporal context obtained with zero motion   leads to significant loss in R-D performance.  These results demonstrate the unequivocal  benefits  of explicit motion estimation. 

 \begin{figure*}[tbp]
  \centering
    \includegraphics[width=0.97\linewidth]{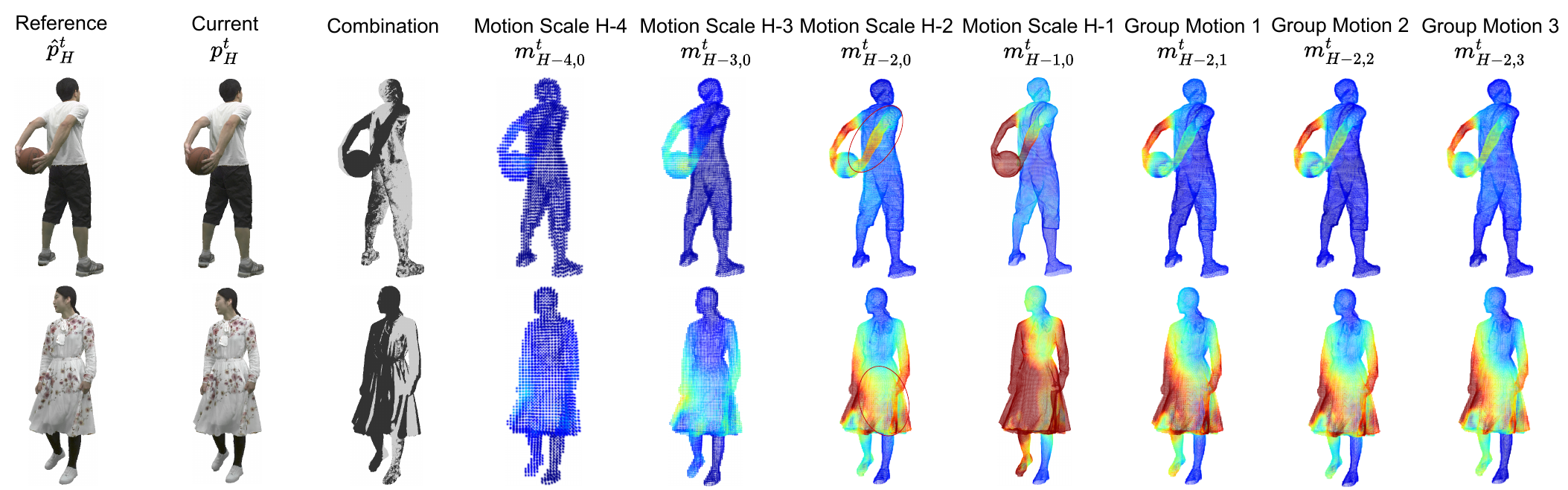}
   \caption{Group motion flow magnitude visualization. In the "Combination" column, gray points belong to reference frame, whereas black points belong to current frame. Warmer  color in the visualization of motion denotes larger movements.} 
   \label{fig:motion_vis}
   \vspace{-2mm}
\end{figure*}

 \paragraph{Top-Down Motion Propagation.} For ``w/o top-down'', we trained a model without  the motion propagation block in Fig. \ref{fig:uinter} (i.e.  $e_l^t=\tilde{e}_l^t$ in Fig. \ref{fig:uinter}). 
 Note that this blocks the motion information propagation from fine to coarse layers, which restricts the compression of motion information into its own scale. We found that {\bf only the lowest scale} contains non-zero motion bitstream ($r_{e,H-K}^t$) with this variation, leading to overall significant performance degradation.

 \paragraph{Bottom-Up Motion Predictive Coding.} In the ``w/o bottom up'' variation, we remove the bottom-up  ``Motion CTX Detach`` and ``Motion CTX Restore`` in Fig. 2 for coding the motion information.  This leads to increased motion bits  and reduced R-D performance for all sequences and rate points.

 \paragraph{Multi-scale Motion Compensation.} 
 For ``single-scale MC'', we only perform motion compensation (MC) at scale $H-2$ (8 for 10 bit point cloud) to generate the temporal context. For other scales, only spatial context is used. We choose scale $H-2$ for MC because this is the same scale as in previous point cloud video compression methods \cite{ijcai2022p126,jiang2023end}. Note that although we only do MC in scale $H-2$, we still do U-structured motion estimation, propagation and coding from scale $H-4$ to $H-2$ to accurately estimate the scale $H-2$ motion.
 We can observe significant gain brought by multi-scale MC due to the rich temporal context brought by multi-scale MC that strengthen the coding of each layer.

 \paragraph{Group Motion.} For  the ``w/o group motion'' variation, we change group motion to one-motion-per-scale (group size=1). We can witness a performance drop due to the decay of expressiveness of the motion flow, especially in sequences with larger movements (model\_vox10).

\subsection{Motion Visualization}

Figure \ref{fig:motion_vis} visualizes the decoded group motion generated by our U-Inter module at different scales. Column 4-7 shows the motion (for group 0) from  scale $H-4$ to scale $H-1$. We can observe that 1) finer-scale motion contains more details than coarser-scale motion; 2) finer-scale motion are more consistent and physically correct than coarser scale for areas with large motion. 
We attribute this to the fact that finer-scale features contains more high-frequency details for correct motion estimation. However, if we  directly estimate the motion at scale $H-1$, we do not get accurate results because 1) the perceptive field of local convolution limits the estimation of large motions; 2) it's inefficient in rate-distortion to compress scale $H-1$'s motion directly. 


For different motion groups, we observe that the network tends to make different predictions among groups in some areas with large movements (red circle). The network uses different motion predictions in these areas to enhance the temporal context, leading to the performance gain compared with single-group motion estimation.




\begin{table}[t]
    \centering
    \small
    \caption{Comparison of Computation Time}
    \begin{tabular}{l|cc|cc|cc}
        \toprule
        Method & \multicolumn{2}{c|}{Ours} & \multicolumn{2}{c|}{Unicorn} & \multicolumn{2}{c}{Ges-TM} \\
        \midrule
        precision& 10 & 11 & 10 & 11 & 10 & 11 \\
        \midrule
        enc time (s) &  1.21   &  6.31   &  0.98   &  3.42   &  10.11   &   27.28  \\
        dec time (s)&  0.82   &  4.75   &  0.67   &  2.56   &  7.18   &  33.81   \\ 
        KNN time (s) &  0.19   &  2.55   &  -   &  -   &  -   &   -  \\
        \bottomrule
    \end{tabular}
    \label{tab:comparison}
    \vspace{-3mm}
\end{table}

\subsection{Complexity}
Table~\ref{tab:comparison} compares the attribute compression complexity for our model, Unicorn and G-PCC-GesTM. U-Motion takes more time than Unicorn (about 25\% more for 10 bit point cloud, about 85\% for 11 bit) mainly due to K-nearest-neighbor (K-NN) search for motion compensation. We use pytorch3d's implementation of K-NN with $O(N^2)$ complexity, leading to large amount of computation on point clouds with $0.7-3$ million points. We can observe in Table~\ref{tab:comparison} that K-NN takes up large amount of the encoding/decoding time. However, the K-NN complexity can be greatly reduced if  K-NN search is restricted to a local region.   The geometry compression complexity is described in the supplementary material.

\section{Conclusion}
We present U-Motion, a PCV codec employing both U-structured temporal prediction and coarse-to-fine spatial prediction.  The key innovation in U-Motion is its U-inter module, combining multi-scale motion estimation,  fine-to-coarse motion propagation and coarse-to-fine  predictive coding of motion,  to enable effective multi-scale motion compensation under rate constraints. We show U-Motion's applicability to  geometry and attribute, achieving significant gains over state-of-the-art methods. Our future work includes multi-frame training to alleviate error-propagation and reducing K-NN complexity for motion compensation.

\clearpage
\setcounter{page}{1}
\maketitlesupplementary
\setcounter{section}{0}

\section{Detailed Architecture of each module}
\begin{figure}[t]
  \centering
   \includegraphics[width=0.99\linewidth]{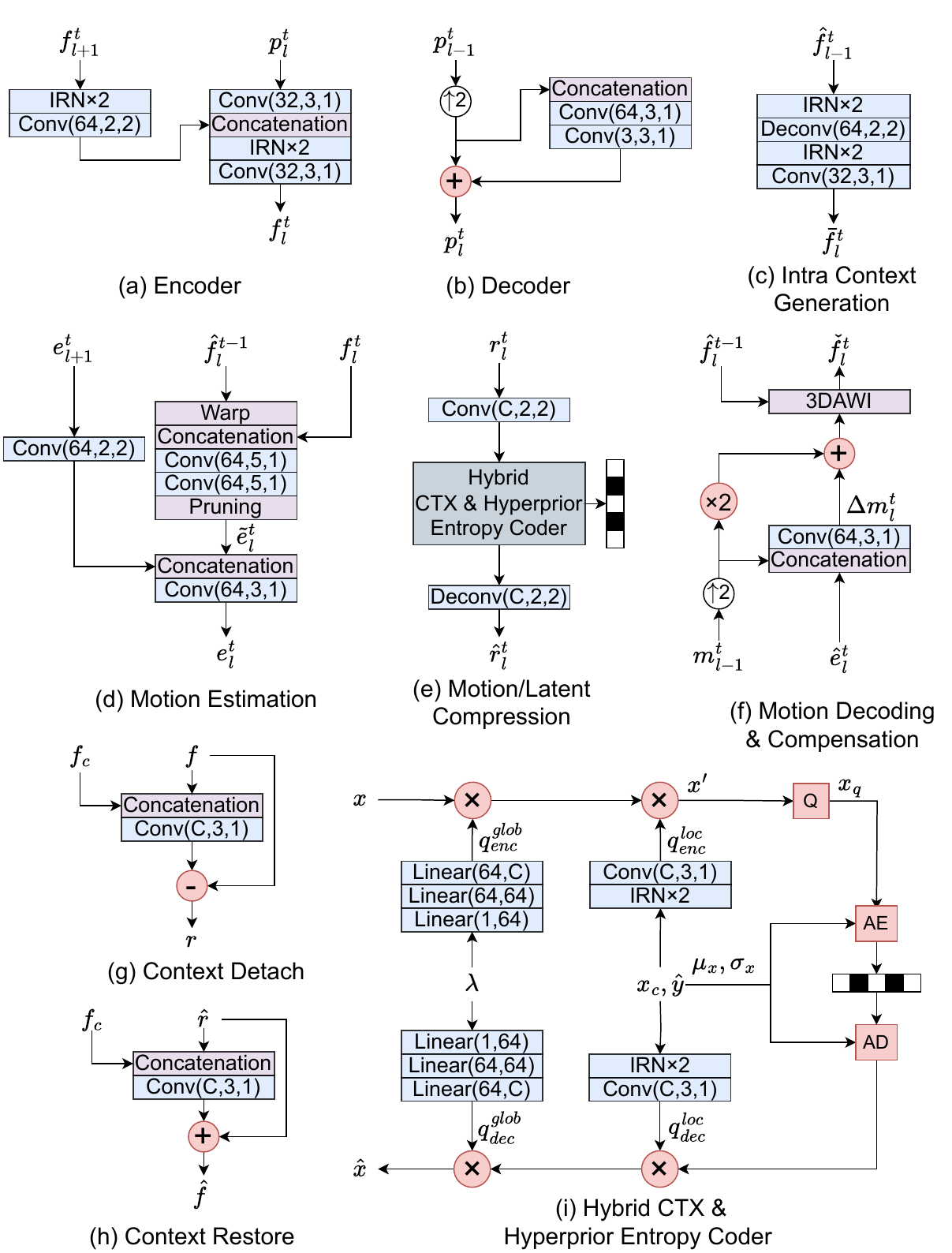}

   \caption{Detailed architectures of each module. Conv($C$, $K$, $S$) denotes a convolution layer with $C$ output channels, $K\times K\times K$ convolution kernel, and $S\times S\times S$ stride. Similar notations is used for Deconv($C$, $K$, $S$), but $S$ indicates the upsampling factor. {\it IRN} denotes the Inception ResNet Block~\cite{wang2021multiscale}. Linear($I$, $O$) denotes a linear layer with $I$ input channels and $O$ output channels. In (e) and (g), $C$ denotes the number of channels of the variable to be entropy coded. $C=32$ for latent compression and $C=24$ for motion compression.}
    \vspace{-2mm}
   \label{fig:detailed}
\end{figure}

This section shows the detailed architecture of each module mentioned in Section \ref{method}. 

\subsection{Motion Estimation}
Note that the concatenation operation in Figure~\ref{fig:detailed} (d) concatenates $\hat{f}_l^{t-1}$ and $f_l^t$, which in general have different sets of geometry coordinates. The concatenation operation of features over two point clouds $p^i$ and $p^j$ is defined over the union of points in $p^i$ and $p^j$  as:

\begin{equation}
f_{cat,u}=\left\{ 
  \begin{array}{l}
  f_u^{i}\oplus f_u^{j}, u\in  p^{i}\cap p^{j}\\
  f_u^{i}\oplus \mathbf{0}, u\in  p^{i},u \notin p^{j}\\
  \mathbf{0}\oplus f_u^{j}, u\in  p^{j},u \notin p^{i}
  \end{array},  \;  u\in  p^{i}\cup p^{j}
\right.
\end{equation}
where $u$ is a coordinate in the union of the geometry coordinates of $f^i$ and $f^j$, $\oplus$ denotes the array concatenation. Note that the concatenation of $\hat{f}_l^{t-1}$ and $f_l^t$ covers more geometry coordinates than $f_l^t$. The pruning operation after convolution  keeps the geometry points in $f_l^t$ only.

\subsection{Motion/Latent Compression}
The architecture of Motion/Residual Compression is shown in Figure~\ref{fig:detailed} (e). The input $e_l^t/r_l^t$ is first downsampled by a stride-two convolution layer, compressed/decompressed by the {\it Hybrid CTX \& Hyperprior Entropy Coder} and upsampled by a stride-two deconvolution layer to the corresponding reconstruction $\hat e_l^t/\hat r_l^t$. $C=32$ for latent compression and $C=24$ for motion compression. The architecture of {\it Hybrid CTX \& Hyperprior Entropy Coder} is shown in Figure~\ref{fig:detailed} (g). We adopt the global-and-local rate control~\cite{li2022hybrid,li2023neural,li2024neural} to achieve various rate points with only one model, controlled by $\lambda$. 
$q_{glob}^{enc}$ and $q_{glob}^{dec}$ are generated from the given $\lambda$ to control the quantization stepsizes for different channels. $q_{loc}^{enc}$ and $q_{loc}^{dec}$ are generated by context $x_c$ and hyperprior $y$ to control local quantization steps. The quantization  of $x$ is formulated by:

\begin{equation}
x_{q,u}=\lfloor x_u\times (1+q_{glob}^{enc}) \times (1+q_{loc,u}^{enc}) \rceil,
\end{equation}
where $u$ is a point in $x$. Rounding is replaced by adding noise during training. The dequantization is formulated by:

\begin{equation}
\hat{x}_u=x_{q,u}\times (1+q_{glob}^{dec}) \times (1+q_{loc,u}^{dec}).
\end{equation}

We denote $x_q$ before quantization as $x'$, and assume that each element of $x'$ follows a Gaussian distribution $N(\mu,\sigma)$, where $\mu$ and $\sigma$ are estimated by the context $x_c$. The probability of $x_q$ can be calculated by: 

\begin{equation}
    p_{x_q}(x_q)=c_{x'}(x_q+0.5)-c_{x'}(x_q-0.5),
\end{equation}
where $c_{x'}$ is the CDF of Gaussian distribution.


\begin{figure*}[htbp]
  \centering
   \includegraphics[width=0.85\linewidth]{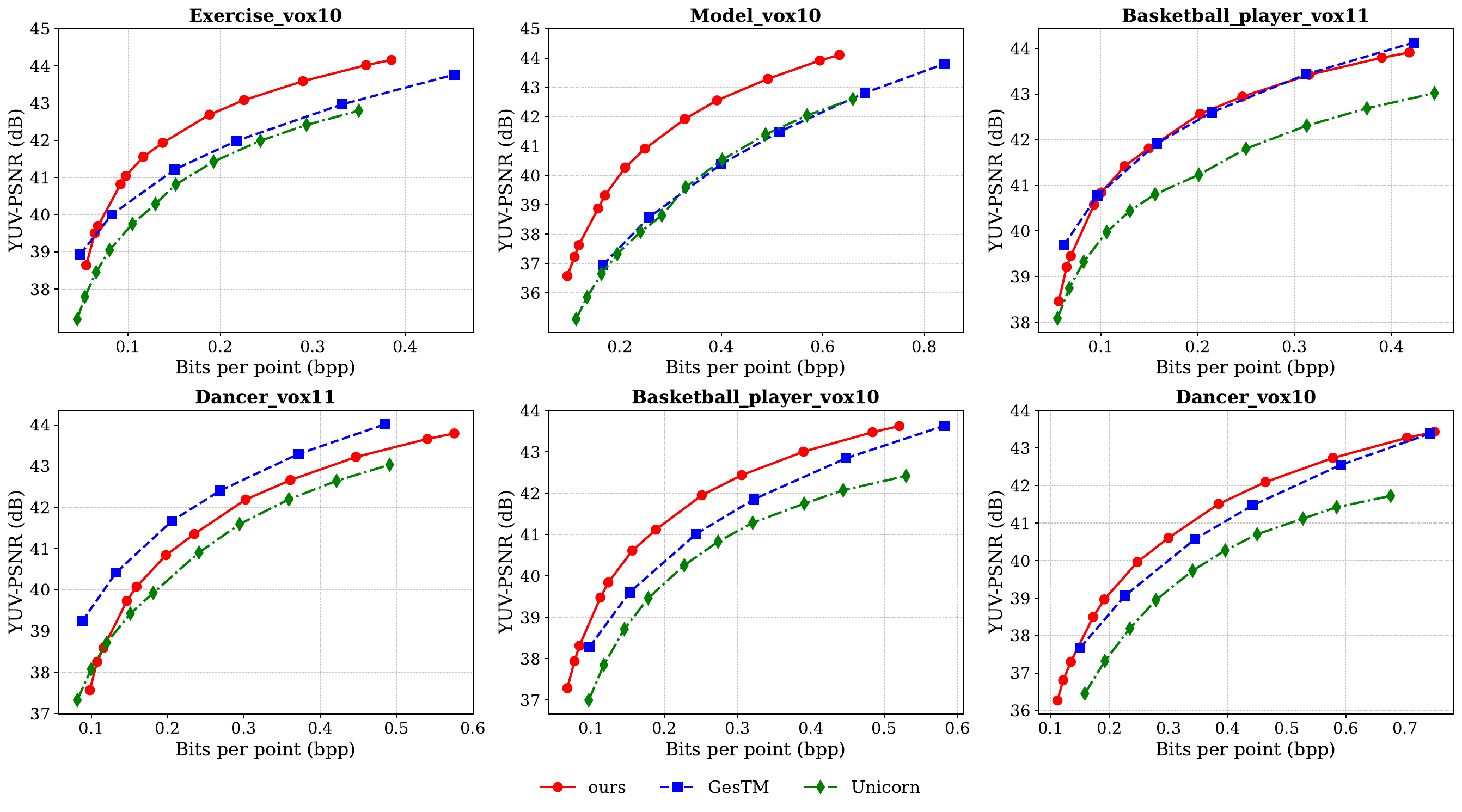}

   \caption{YUV-PSNR Performance comparison on attribute (color) compression among our method, Unicorn and G-PCC-GesTM.}
   \vspace{-2mm}
   \label{fig:res_attr_yuv}
\end{figure*}

\begin{figure*}[t]
  \centering
   \includegraphics[width=0.92\linewidth]{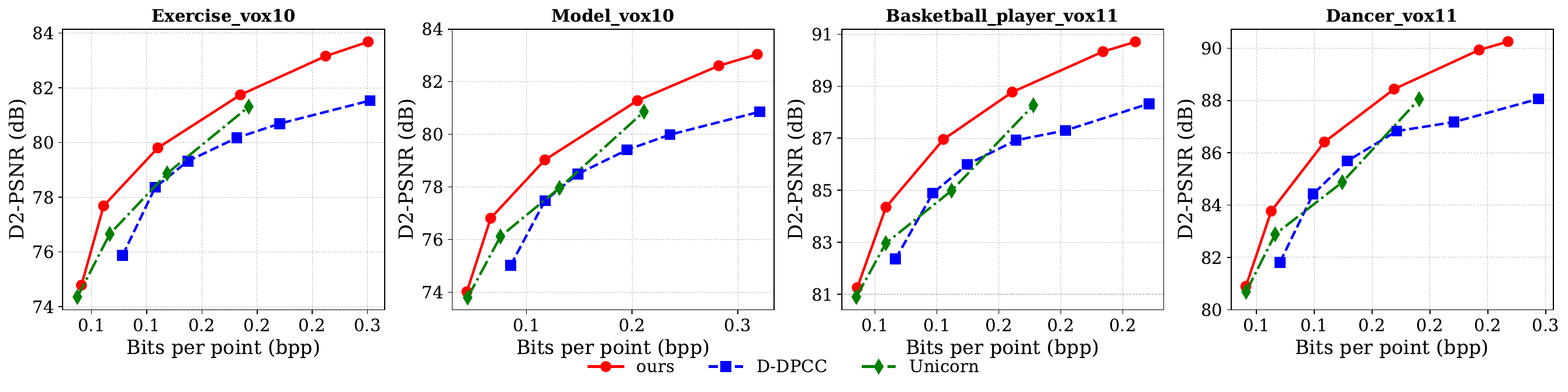}

   \caption{D2-PSNR performance comparison on lossy geometry compression among our method, Unicorn and D-DPCC}
   \label{fig:res_geo_d2}
   \vspace{-2mm}
\end{figure*}

\section{Experiments}
\subsection{Settings for Attribute Compression}
Our U-Motion attribute compression model consists of five layers, {\it i.e.} five U-Inter layer and five Spatial-Temporal Predictive Coding layer. We adopt the base layer strategy employed in YOGA~\cite{zhang2023yoga}, which compresses the lower-scale point cloud $p_{H-3}^t$ with G-PCC-GesTM to provide a base for the hierarchical reconstruction. Note that for layer $l\in\left[L,H-3\right]$, although the color is coded by GesTM, there still exists the motion embedding $e_l^t$ and latent residual $r_l^t$ to be coded by our network for higher layers' reconstruction. We do not losslessly compress $p_l^t$ at layer $H-3$ due to the excessive bit-rate consumption. Instead, we use different GesTM quantization parameters ($QP$) for different rate points to achieve the R-D balance of overall performance. We traverse different combinations of $\lambda$ and $QP$ on a subset of dataset and get the most R-D efficient pairs: $\lambda=$[300, 460, 705, 910, 1655, 2537, 3888, 5960, 9134, 14000, 16000], and the corresponding $QP=$[20, 20, 16, 16, 16, 16, 12, 12, 8, 8, 8].

\subsection{Settings for Geometry Compression}
As mentioned in Section~\ref{geom}, the APU module~\cite{wang2024versatile1} supports both lossless and lossy geometry compression. For geometry, we use different combinations of lossless/lossy compression layers and different $\lambda$ to realize different rate points. Given a geometry U-Motion model with $(H-L+1)$ layers and $\lambda$ as R-D factor, where $L\rightarrow M$ layers are lossless and $M+1\rightarrow H$ are lossy, the loss function $\cal L$ during training is:

\begin{equation}
    {\cal L}=\sum_{l=L}^H R_{r,l}+R_{m,l}+\lambda_lBCE(\tilde{p}_l^t, p_l^t),
\end{equation}
where $\lambda_l=1$ for lossless layer, $\lambda_l=\lambda$ for lossy layer, $\tilde{p}_l^t$ is the decoded occupancy probability mentioned in Section~\ref{geom}. We train six models corresponding to the six rate points. The first three models consist of 1 lossless layer and 2 lossy layer, with $\lambda=0.5,1,2$. The last three models consist of 2 lossless layers and 1 lossy layer, with $\lambda=2,5,7$. 

\begin{table*}[t]
    \centering
    \small
     \caption{the bpp and Y-PSNR of the visualized point cloud videos.}
    \begin{tabular}{cc cc cc cc}
      \toprule
      \multicolumn{2}{c}{} & \multicolumn{2}{c}{qp1} & \multicolumn{2}{c}{qp2} & \multicolumn{2}{c}{qp3} \\ 
      \multicolumn{2}{c}{ } & bpp & Y-PSNR & bpp & Y-PSNR & bpp & Y-PSNR\\ 
      \midrule
      \multicolumn{2}{c}{exercise\_vox10} & 0.071  & 37.09  & 0.257  & 41.53 & 0.479 & 42.87\\
      \multicolumn{2}{c}{model\_vox10} & 0.098  & 36.29  & 0.329  & 41.19 & 0.742 & 43.02\\
      \multicolumn{2}{c}{basketball\_vox11} & 0.081  & 36.62  & 0.291  & 41.42 & 0.425 & 42.74\\
      \multicolumn{2}{c}{dancer\_vox11} & 0.142  & 34.77  & 0.474  & 41.13 & 0.528 & 42.50\\
      \bottomrule
     \end{tabular}
      \label{tab:bd_visualize}
\end{table*}  



\subsection{Results}
\paragraph{YUV-PSNR for Attribute Compression.} We provide the YUV R-D curve in Figure~\ref{fig:res_attr_yuv}. Note that the MPEG reference software {\it pc\_error} does not compute YUV-PSNR directly. Instead, we use {\it pc\_error} to get Y-, U- and V-PSNR and calculate YUV-PSNR as:
\begin{equation}
    PSNR_{YUV}=\frac{6 PSNR_Y+PSNR_U+PSNR_V}{8}.
\end{equation}
We can observe that the performance comparison is basically consistent with Y-PSNR result.

\paragraph{D2-PSNR for Geometry Compression.} We provide the D2 R-D curve in Figure~\ref{fig:res_geo_d2}. We can observe that the performance comparison is basically consistent with D1-PSNR result.

\begin{figure}[t]
  \centering
   \includegraphics[width=0.99\linewidth]{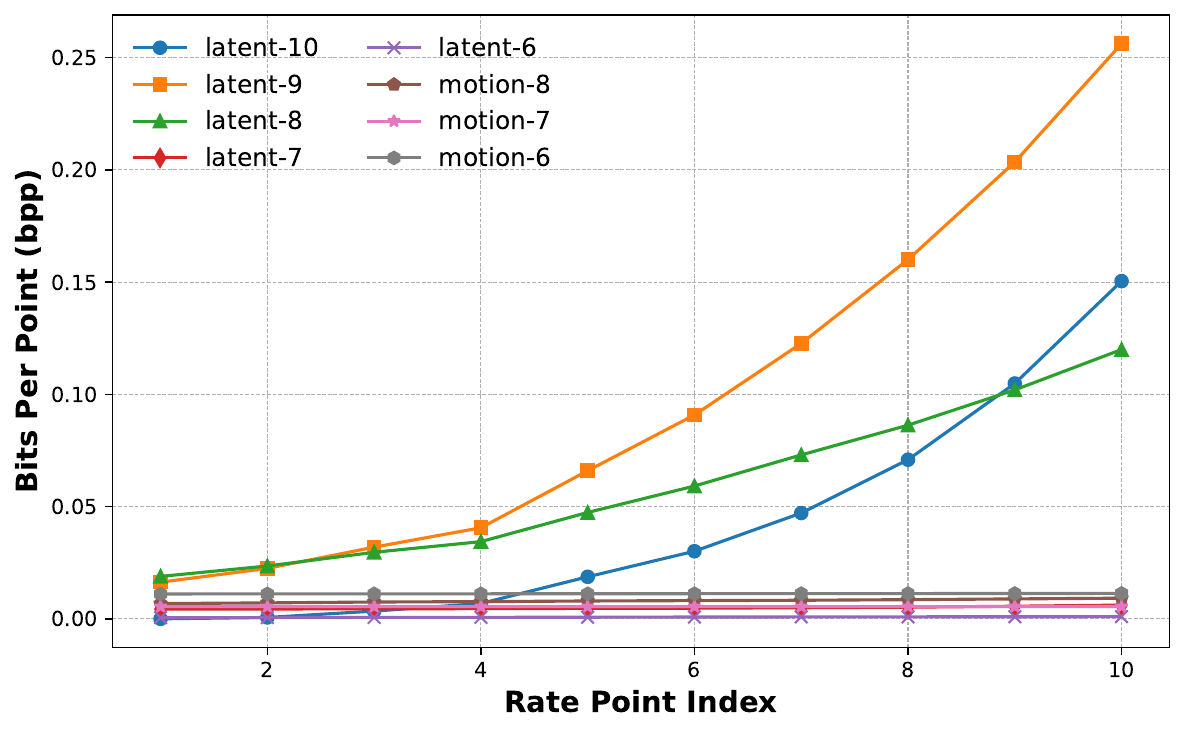}

   \caption{Rate allocation between motion and latent over different layers corresponding to different rate points for {\it model}, frame index 00000150.}
    \vspace{-2mm}
   \label{fig:rates}
\end{figure}

\paragraph{Rate Allocation.} Figure~\ref{fig:rates} shows the bit-rate for coding each layer's motion embedding $e_l^t$ and latent residual $r_l^t$ for one test sequence. The same trend is observed for other sequences. {\it latent-$l$} and {\it motion-$l$} denotes the bit-rate consumption of coding $r_{l}^t$ and $e_{l}^t$. Note that layers not plotted have zero bits. We can observe that: 1) The bit-rate of motion information remains consistent regardless of the overall bit-rate. We attribute this to the fact that motion bit-rate only contributes to a small portion of overall bit-rate, and reducing motion bit-rate may result in the low quality of inter context $\check{f}_l^t$. Therefore, the learnt network may have used the same quantization stepsizes regardless the target total rate.  2) Only the two lowest layers (Layer 6 and 7) have  motion bits. We believe that this is  due to the inefficiency of coding residual motion in higher layers, relative to using all the bits for coding the latent. 3) The rate allocation for the latents across layers depend on the target total bit rate. At the higher rate, the higher layers take more bits (except the highest layer).  At the lower rate, some intermediate layers were given more rates.  Given these observations, in future improved versions of the codec, we may choose not to estimate residual motion at higher layers, and directly use the upsampled motion from lower layers for generating the inter context at higher layers. This may reduce the encoding complexity, without sacrificing the rate-quality trade-off.   

\begin{table}[t]
    \centering
    \small
    \caption{Comparison of Geometry Computation Time}
    \begin{tabular}{l|cc|cc|cc}
        \toprule
        Method & \multicolumn{2}{c|}{Ours} & \multicolumn{2}{c|}{Unicorn} & \multicolumn{2}{c}{D-DPCC} \\
        \midrule
        precision& 10 & 11 & 10 & 11 & 10 & 11 \\
        \midrule
        enc time (s) &  0.64   &  7.42   &  0.45   &  3.15   &  0.67   &   8.75  \\
        dec time (s)&  0.56   &  6.95   &  0.52   &  3.42   &  0.56   &  8.42   \\
        \bottomrule
    \end{tabular}
    \label{tab:geo_comparison}
\end{table}

\paragraph{Geometry Compression Complexity.} For geometry compression, the complexity is related to the model architecture of each rate point. We only compare the model related to the highest rate point, {\it i.e.} 1 lossy and 2 lossless layers. The complexity is shown in Table~\ref{tab:geo_comparison}.

\paragraph{Visualization of Motion Video.} We visualize the point cloud motion magnitude video in folder {\it motion\_mag}. We can observe that areas with larger movements have brighter colors.

\paragraph{Visualization of Reconstructed Video.} We visualize the reconstructed point cloud video in folder {\it recon}. We pick one high bit-rate ($\lambda=16000,QP=8$), one medium bit-rate ($\lambda=5960, QP=12$) and one low bit-rate ($\lambda=300, QP=20$). The corresponding bpp and Y-PSNR of each sequence, each rate point is shown in Table~\ref{tab:bd_visualize}.

{
    \small
    \bibliographystyle{ieeenat_fullname}
    \bibliography{main}
}

\end{document}